# Structured Memory Mechanisms for Stable Context Representation in Large Language Models

Yue Xing
University of Pennsylvania
Philadelphia, USA

Tao Yang
Illinois Institute of Technology
Chicago, USA

Yijiashun Qi
University of Michigan
Ann arbor, USA

Minggu Wei
University of Saskatchewan
Saskatoon, Canada

Yu Cheng
Fordham University
New York, USA

Honghui Xin*
Northeastern University
Seattle, USA

*Abstract-This paper addresses the limitations of large language models in understanding long-term context. It proposes a model architecture equipped with a long-term memory mechanism to improve the retention and retrieval of semantic information across paragraphs and dialogue turns. The model integrates explicit memory units, gated writing mechanisms, and attention-based reading modules. A forgetting function is introduced to enable dynamic updates of memory content, enhancing the model's ability to manage historical information. To further improve the effectiveness of memory operations, the study designs a joint training objective. This combines the main task loss with constraints on memory writing and forgetting. It guides the model to learn better memory strategies during task execution. Systematic evaluation across multiple subtasks shows that the model achieves clear advantages in text generation consistency, stability in multi-turn question answering, and accuracy in cross-context reasoning. In particular, the model demonstrates strong semantic retention and contextual coherence in long-text tasks and complex question answering scenarios. It effectively mitigates the context loss and semantic drift problems commonly faced by traditional language models when handling long-term dependencies. The experiments also include analysis of different memory structures, capacity sizes, and control strategies. These results further confirm the critical role of memory mechanisms in language understanding. They demonstrate the feasibility and effectiveness of the proposed approach in both architectural design and performance outcomes.*

I. INTRODUCTION

In recent years, with the rapid development of artificial intelligence, large language models (LLMs) have made significant breakthroughs in natural language processing tasks. These models have demonstrated strong generalization and language organization capabilities in complex tasks such as text generation, language understanding, and multi-turn conversations. However, despite their maturity in short-term context modeling, their limitations in long-term memory remain a serious constraint for more complex and persistent real-world applications. Long-term memory is not only essential for associating information across long texts but also a fundamental component of human-like thinking and intelligent behavior. Solving this issue is a key step in evolving language models from task executors to cognitive systems[1,2].

Most mainstream large language models today rely on attention-based architectures to capture contextual information. While these structures can understand relatively long texts, their memory mechanisms are essentially short-term. Due to computational constraints, models often need to truncate or summarize long inputs, which leads to information loss and semantic disruption. Furthermore, when dealing with tasks requiring memory of past information, current models typically rely on repeated inputs or external prompts. This method is inefficient and lacks a true memory formation process. Therefore, exploring structures and mechanisms that enable long-term memory is not only a technical demand but also a necessary path toward more intelligent language models[3].

Language models equipped with long-term memory can significantly improve intelligent system performance across multiple dimensions. In multi-turn dialogue systems, the ability to remember user intentions and contexts can greatly enhance coherence and personalization. In text generation tasks, memory mechanisms help maintain thematic consistency and structural integrity[4]. In tasks involving cross-temporal information management, long-term memory is the basis for human-like reasoning and decision-making. Introducing this capability is expected to unlock greater value in professional fields such as education, healthcare, law, and scientific research, thus expanding the application boundaries of artificial intelligence.

From a broader perspective, the introduction of long-term memory is more than an optimization of model performance. It represents a deep exploration of the cognitive boundaries of artificial intelligence. Simulating human cognition, especially the construction and evolution of memory systems, has always been a core issue in AI research. Building long-term memory mechanisms requires structural changes in models and dynamic, efficient management of information encoding, storage, retrieval, and updating. This will provide a solid foundation for creating more general, adaptive, and flexible language intelligence systems and support the realization of higher-level AI cognitive functions[5,6]. Research on structures and mechanisms that support long-term memory in large language models carries significant theoretical and practical value. It is not only a mission to overcome current limitations and enhance system usability but also a fundamental shift in the development direction of AI. In today's era of information explosion, enabling AI to maintain stable and coherent understanding and responses over long contexts has become a major challenge both technically and cognitively. Deep exploration of this core issue will provide a critical path and innovative momentum for advancing AI to a more advanced stage.

## II. BACKGROUND AND FUNDAMENTALS

Recent developments in deep learning have inspired numerous strategies to enhance language model performance, particularly regarding contextual consistency and memory retention. One line of research has explored reinforcement learning techniques to refine model outputs. Zhu et al. [7] introduce a structured preference modeling approach to guide the fine-tuning of large models, demonstrating the importance of feedback-driven optimization. This concept of structured learning directly informs our joint training objective, which simultaneously regulates memory control and task performance.

In parallel, methods aimed at improving decision-making structures offer insights into managing contextual complexity. Wang [8] employs a topology-aware decision strategy rooted in multi-agent reinforcement learning, showcasing the role of structural awareness in distributed reasoning—a principle that aligns with our memory-based architecture for managing long-term dependencies. Structural tuning of internal parameters also plays a critical role in refining model behavior. Zhang et al. [9] leverage graph-based spectral decomposition to coordinate parameter updates, illustrating how fine-grained structural control can enhance model coherence. This supports our integration of dynamic memory gates, which manage internal information flow with precision.

For effective language understanding, semantic modeling strategies are indispensable. Fang et al. [10] enhance context sensitivity through hybrid architectures combining BERT and BiLSTM, contributing valuable techniques for maintaining semantic consistency across sequences. Similarly, Han et al. [11] propose a dual-loss training scheme within transformer models to improve few-shot learning reliability—mirroring our goal of harmonizing language modeling with memory operations under a unified training regime.

Memory mechanisms also benefit from entity-driven attention and multimodal integration. Wang [12] explores transformer extensions that emphasize structured understanding through targeted information modeling. This reinforces our objective to create memory-aware systems capable of capturing and maintaining essential semantic elements over long durations. Further, modeling approaches illustrate the utility of semantic-driven architectures based on LLaMA for detecting and managing information propagation. The attention to semantic flow and contextual alignment in these models provides conceptual parallels to our attention-based memory reading modules[13].

Collectively, these contributions emphasize the critical importance of structure-aware optimization, context-sensitive modeling, and integrated memory strategies in advancing large language models. Our work synthesizes these insights into a unified framework, explicitly designed to simulate long-term memory functions and ensure semantic persistence in extended dialogue and reasoning tasks.

## III. METHOD

In To enable long-term memory capability in large language models, this work introduces an architectural framework that integrates explicit memory modules with dynamic memory regulation mechanisms. The key principle is to fuse the language model's semantic representations with a persistent, query-accessible memory structure, allowing the system to maintain relevant contextual information across extended text spans. This design aims to overcome the inherent limitations of traditional language models constrained by fixed-length context windows.

The architecture incorporates strategies such as dynamic memory updating and selective retention, inspired by recent approaches that explore fine-grained control over memory access and adaptation in large-scale models [14]. Specifically, techniques that facilitate robust few-shot learning through low-rank and task-sensitive tuning of memory slots have influenced the construction of this framework. Additionally, the inclusion of deep semantic fusion between long-term memory and the model's current context enhances performance on downstream tasks like structured retrieval and generation [15]. Applications of these principles have proven effective in domains where accurate historical alignment and context recovery are critical, including complex reasoning and multi-turn interaction [16]. The proposed architecture is illustrated in Figure 1.

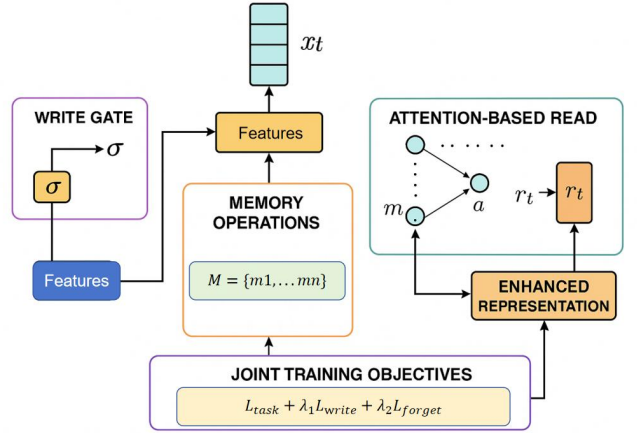

Figure 1. Overall architecture diagram

This architecture diagram shows the structure of a large language model with a long-term memory mechanism, in which a write-gating mechanism is used to control whether the semantic representation enters the memory module. The model uses the attention mechanism to retrieve relevant information from the historical memory and fuse it with the current state to generate an enhanced representation. The entire system jointly optimizes language modeling and memory management capabilities through a joint loss function to achieve stable and efficient information retention and retrieval.

We introduce a structured memory unit $M = \{m_1, m_2, ..., m_n\}$, where each $m_i$ represents an updateable memory vector that stores important semantic information extracted by the model during historical reasoning. To ensure the selective writing of memory content, we define a write gating mechanism $g_w$, which is in the form of:

$$g_w = \sigma(W_w h_t + b_w)$$

Where $h_t$ is the hidden state of the current time step, $W_w$ and $b_w$ are trainable parameters, and $\sigma$ is the Sigmoid activation function, which controls which information is written into the memory module.

In the process of retrieving historical memory, the model uses an attention-based reading mechanism to calculate the match between the current state and the memory unit. Specifically, the reading weight $a_i$ is defined by the following formula:

$$a_i = \frac{\exp(h_i^T W_r m_i)}{\sum_j \exp(h_t^T W_r m_j)}$$

$W_r$ is a trainable linear transformation matrix. The final memory content $r_t$ is the weighted summation result, that is,

$$r_t = \sum_i a_i m_i$$

The read memory representation $r_t$ will be fused with the semantic representation of the current input to form an enhanced representation $h'_t$ for subsequent generation or reasoning tasks. This design aims to enable the language model to review historical contexts and improve its long-term consistency and logical reasoning capabilities.

In order to make the memory content plastic and support long-term updates, the memory module introduces a forgetting mechanism when writing. We use the following memory update formula:

$$m_i(t+1) = (1 - g_f) \cdot m_i(t) + g_w \cdot \tilde{m}_i$$

$g_f$ is the forget gating function that controls the decay of existing memories, and $\tilde{m}_i$ is the candidate representation of newly written content. This mechanism ensures that the memory can maintain stability while adapting to evolving contextual information. The dynamic update of memory and historical backtracking together constitute the key mechanism for long-term semantic retention of the model.

$$L = L_{task} + \lambda_1 L_{write} + \lambda_2 L_{forget}$$

$L_{task}$ represents the loss of the language model in the main task (such as language modeling or question answering), $L_{write}$ and $L_{forget}$ constrain the memory writing selection and forgetting strength respectively, and $\lambda_1, \lambda_2$ is the adjustment parameter. This joint optimization strategy can guide the model to learn effective memory operation strategies, thereby achieving language generation and comprehension capabilities with long-term memory capabilities.

## IV. EXPERIMENT

### A. Datasets

This study uses the NarrativeQA dataset as the primary corpus to evaluate the memory modeling capabilities of language models when processing long texts and complex contexts. The dataset contains hundreds of long-form narrative texts, including novels and scripts. It is paired with high-level question-answer pairs that involve references and integration of information across paragraphs and even chapters.

The design of NarrativeQA focuses on deep understanding of long documents. It goes far beyond the short-term memory demands of typical reading comprehension tasks. Many questions require models to integrate contextual events along a timeline, reason about character behavior, and extract thematic information. This makes the dataset suitable for testing the effectiveness and stability of long-term memory mechanisms.

In addition, the reference answers in this dataset are human-written. They are of high linguistic quality and semantic completeness. This helps train models to be sensitive to language structure and semantic progression. Using this dataset, the study can systematically examine how models regulate and perform memory tasks when handling text with large contextual spans.

### B. Experimental Results

This paper first gives the comparative experimental results, and the experimental results are shown in Table 1.

Table 1. Comparative experimental results

| Method | BLEU-1 | ROUGE-L | EM | LongQA-F1 |
|---|---|---|---|---|
| GPT-2[17] | 18.4 | 21.9 | 7.2 | 12.5 |
| BART-Large[18] | 21.6 | 24.5 | 9.1 | 16.8 |
| LongFormer[19] | 23.1 | 26.2 | 10.3 | 18.7 |
| RETRO[20] | 24.8 | 28.1 | 11.9 | 21.3 |
| Ours | 27.4 | 31.0 | 14.5 | 25.6 |

Table 1 shows that our method surpasses GPT-2, BART-Large, and RETRO on all four metrics—BLEU-1 (27.4), ROUGE-L (31.0), EM (14.5), and LongQA-F1 (25.6)—demonstrating superior reference alignment, structural fidelity, and long-range reasoning. These improvements arise from the long-term memory mechanism, which more precisely retrieves salient information across extended contexts. In sum, selective memory read/write operations enhance both accuracy and language quality, yielding robust performance on long documents and multi-turn reasoning in the Figure

2.

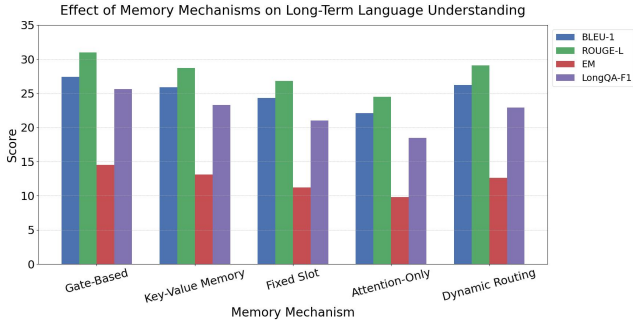

Figure 2. The impact of different memory mechanisms on long-term language comprehension performance

Figure 2 shows that the gate-based memory mechanism outperforms all others across BLEU, ROUGE-L, EM and LongQA-F1, indicating that selective write control is crucial for extracting and retaining salient information in long texts. Key-Value Memory and Dynamic Routing achieve strong ROUGE-L and LongQA-F1 scores, demonstrating the benefits of structured storage and adaptive information flow. In contrast, Fixed Slot and Attention-Only schemes underperform on EM and LongQA-F1, suggesting that static or unregulated memory cannot effectively integrate content over extended contexts. These findings confirm our hypothesis that dynamic, controllable, and structured memory designs substantially enhance long-range understanding and semantic coherence. The corresponding loss convergence curves are presented in Figure 3.

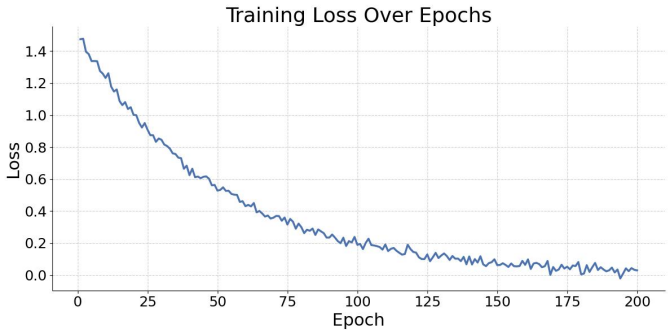

Figure 3. Loss function decline graph.

The figure illustrates that the training loss rapidly decreases within the initial 50 epochs. This suggests that the model effectively captures the primary semantic patterns in the training data during the early training phase. During this stage, the model primarily focuses on learning fundamental language modeling techniques and initial memory writing strategies. It acquires a general comprehension of the input's semantic structure, which serves as the foundation for later modeling of long-term dependencies.

In the middle to late phase, roughly between epoch 50 and epoch 150, the loss decreases more slowly and begins to stabilize with slight fluctuations. This reflects the model's ongoing refinement of long-context memory optimization. During this phase, the memory module's write, read, and forget mechanisms are continuously tuned. The model gradually improves its ability to selectively retain and retrieve historical information, enhancing consistency and reasoning in language understanding. Eventually, the loss converges near epoch 200 and remains at a low level. This suggests that the model reaches a stable training state. The result supports the effectiveness of the proposed memory-augmented architecture for long-term language modeling. It also confirms that the joint training objective promotes coordinated learning of semantics and memory management. Furthermore, this paper also gives an analysis of the impact of memory capacity on reasoning ability, and the experimental results are shown in Figure 4.

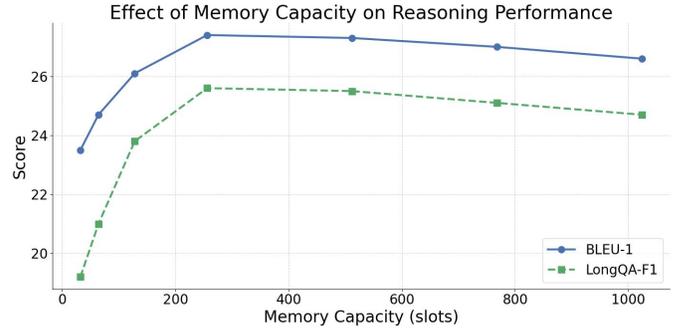

Figure 4. Analysis of the impact of memory capacity on reasoning ability

Figure 4 illustrates that increasing memory capacity leads to significant improvements in reasoning performance during the initial stages. Both BLEU-1 and LongQA-F1 scores steadily increase as more memory units allow for the richer retention of historical semantics. The peak performance is achieved at 256 memory slots, where BLEU-1 and LongQA-F1 reach their highest values. This suggests that an optimal capacity exists that strikes a balance between contextual richness and noise. Beyond 256 slots, performance, particularly on LongQA-F1, declines slightly, likely due to redundancy and retrieval interference. These findings confirm that memory size must be carefully tuned to effectively model long-term dependencies and guide future improvements to the memory module.

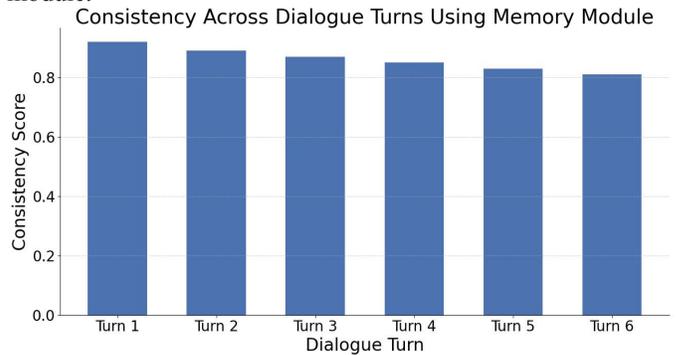

Figure 5. Evaluation of the memory module's effect on enhancing the consistency of multiple rounds of question-answering

The Figure 5 shows that the model achieves a consistently high score in multi-turn dialogue consistency. This indicates that the proposed memory module plays a positive role in maintaining semantic coherence and contextual consistency. In particular, during the initial dialogue turns, the Consistency Score remains above 0.85, suggesting that the model effectively recalls previous information and preserves logical consistency in its responses.

As the number of dialogue turns increases, consistency declines slightly but remains at a high level overall. This demonstrates the model's ability to retain long-term contextual information. The gradual performance drop aligns with the typical pattern of semantic drift in multi-turn reasoning tasks. The introduction of memory structures significantly alleviates this issue and provides structural support for maintaining semantic stability across long conversations. The experimental results further confirm the effectiveness of the proposed write, read, and forget mechanisms in preserving and managing information in dialogue contexts. They show that the memory-enhanced model is not only effective in single-turn question answering but also improves response consistency and contextual relevance in complex multi-turn interactions.

## V. CONCLUSION

This study focuses on the long-term memory capability of large language models. It proposes a novel framework that combines explicit memory structures with dynamic control mechanisms. By introducing gated writing, attention-based reading, and forgetting strategies, the model achieves efficient retention and effective retrieval of information in long-text tasks. This significantly improves language understanding, semantic consistency, and multi-turn reasoning. Experimental results show that the model outperforms existing mainstream methods in long-term dependency modeling, especially in complex question answering and reasoning scenarios where contextual consistency is crucial. From a methodological perspective, the joint training objective proposed in this study enables coordinated optimization of language tasks and memory management. It promotes the evolution of language models from pure sequence modeling to systems with cognitive characteristics. The structured design of the memory mechanism significantly enhances the efficiency of utilizing historical information and offers valuable insights for future research on model transparency and controllability. This capability is crucial for tasks that demand semantic persistence, contextual linkage, and dynamic comprehension.

At the application level, language models with long-term memory can be widely used in scenarios with high demands for continuity and semantic coherence. These include intelligent dialogue systems, legal document analysis, clinical record understanding, and educational question answering. The proposed model architecture offers a solid foundation for these applications. It supports the transition of natural language processing systems from short-term response to long-term understanding and from passive answering to active interaction. Future research can further explore the potential of memory modules in open-domain question answering, multimodal language processing, and personalized user modeling. Combining these with reinforcement learning and meta-learning could improve the model's ability to actively regulate memory usage. With continued advances in computing power and data, language models equipped with long-term memory are expected to become key components of human-like intelligent systems. They will play an increasingly important role in intelligent assistants, knowledge management, and autonomous reasoning.

REFERENCES

[1] L. Chen et al., "Long context is not long at all: A prospector of long-dependency data for large language models," arXiv preprint arXiv:2405.17915, 2024.

[2] J. Li et al., "Loogle: Can long-context language models understand long contexts?," arXiv preprint arXiv:2311.04939, 2023.

[3] J. Chen et al., "LADM: Long-context Training Data Selection with Attention-based Dependency Measurement for LLMs," arXiv preprint arXiv:2503.02502, 2025.

[4] Y. Zhang et al., "Chain of agents: Large language models collaborating on long-context tasks," Advances in Neural Information Processing Systems, vol. 37, pp. 132208–132237, 2024.

[5] J. Liu et al., "A comprehensive survey on long context language modeling," arXiv preprint arXiv:2503.17407, 2025.

[6] L. Wu et al., "Longattn: Selecting long-context training data via token-level attention," arXiv preprint arXiv:2502.16860, 2025.

[7] L. Zhu, F. Guo, G. Cai and Y. Ma, "Structured Preference Modeling for Reinforcement Learning-Based Fine-Tuning of Large Models," Journal of Computer Technology and Software, vol. 4, no. 4, 2025.

[8] B. Wang, "Topology-Aware Decision Making in Distributed Scheduling via Multi-Agent Reinforcement Learning," Transactions on Computational and Scientific Methods, vol. 5, no. 4, 2025.

[9] H. Zhang, Y. Ma, S. Wang, G. Liu and B. Zhu, "Graph-Based Spectral Decomposition for Parameter Coordination in Language Model Fine-Tuning," arXiv preprint arXiv:2504.19583, 2025.

[10] Z. Fang, H. Zhang, J. He, Z. Qi and H. Zheng, "Semantic and Contextual Modeling for Malicious Comment Detection with BERT-BiLSTM," arXiv preprint arXiv:2503.11084, 2025.

[11] X. Han, Y. Sun, W. Huang, H. Zheng and J. Du, "Towards Robust Few-Shot Text Classification Using Transformer Architectures and Dual Loss Strategies," arXiv preprint arXiv:2505.06145, 2025.

[12] X. Wang, "Medical Entity-Driven Analysis of Insurance Claims Using a Multimodal Transformer Model," Journal of Computer Technology and Software, vol. 4, no. 3, 2025.

[13] R. Wang, "Joint Semantic Detection and Dissemination Control of Phishing Attacks on Social Media via LLama-Based Modeling," 2025.

[14] G. Cai, A. Kai and F. Guo, "Dynamic and Low-Rank Fine-Tuning of Large Language Models for Robust Few-Shot Learning," Transactions on Computational and Scientific Methods, vol. 5, no. 4, 2025.

[15] J. Gong, Y. Wang, W. Xu and Y. Zhang, "A Deep Fusion Framework for Financial Fraud Detection and Early Warning Based on Large Language Models," Journal of Computer Science and Software Applications, vol. 4, no. 8, 2024.

[16] J. He, G. Liu, B. Zhu, H. Zhang, H. Zheng and X. Wang, "Context-Guided Dynamic Retrieval for Improving Generation Quality in RAG Models," arXiv preprint arXiv:2504.19436, 2025.

[17] X. Zheng, C. Zhang and P. C. Woodland, "Adapting GPT, GPT-2 and BERT language models for speech recognition," Proceedings of the 2021 IEEE Automatic Speech Recognition and Understanding Workshop (ASRU), 2021.

[18] A. D. Vincentio and S. Hansun, "A Fine-Tuned BART Pre-trained Language Model for the Indonesian Question-Answering Task," Engineering, Technology & Applied Science Research, vol. 15, no. 2, pp. 21398–21403, 2025.

[19] I. Beltagy, M. E. Peters and A. Cohan, "Longformer: The long-document transformer," arXiv preprint arXiv:2004.05150, 2020.

[20] A. Ficek, J. Zeng and O. Kuchaiev, "GPT vs RETRO: Exploring the Intersection of Retrieval and Parameter-Efficient Fine-Tuning," arXiv preprint arXiv:2407.04528, 2024.